\documentclass[runningheads]{llncs}
\usepackage{graphicx}
\usepackage{hyperref}
\usepackage{url}
\usepackage{array} 

\usepackage{algorithm}
\usepackage{algpseudocode}
\usepackage{amsmath} 

\usepackage{multirow}

\usepackage{pgfplots}
\pgfplotsset{compat=1.17}

\usepackage{caption}

\usepackage{cite}
\usepackage{booktabs}
\usepackage{makecell}
\usepackage{bbm}
\usepackage{amssymb}

\newcommand{\Output}[1]{\item[\textbf{Output:}] #1}
\newcommand{\Input}[1]{\item[\textbf{Input:}] #1}

\usepackage[misc]{ifsym}

\begin{document}
\title{MATEval: A Multi-Agent Discussion Framework for Advancing Open-Ended Text Evaluation}
\titlerunning{MATEval: A Multi-Agent Text Evaluation Framework}
%
\author{Yu Li\inst{1,}\thanks{Equal Contributors.}\and
Shenyu Zhang\inst{1,\star} \and
Rui Wu\inst{2} \and
Xiutian Huang\inst{2} \and
Yongrui Chen\inst{1} \and \\
Wenhao Xu\inst{2}\textsuperscript{(\Letter)} \and
Guilin Qi\inst{1} \textsuperscript{(\Letter)}\and
Dehai Min\inst{1}}
\authorrunning{Y. Li et al.}
%
\institute{Southeast University, Nanjing, China\\
\email{\{yuli\_11, shenyuzhang, yrchen, gqi, zhishanq\}@seu.edu.cn}\\
\and
Ant Group, Hangzhou, China\\
\email{\{guli.wr, xiutian.hxt, hao.xuwh\}@antgroup.com}}
\maketitle              

\begin{abstract}

Recent advancements in generative Large Language Models (LLMs) have been remarkable, however, the quality of the text generated by these models often reveals persistent issues. Evaluating the quality of text generated by these models, especially in open-ended text, has consistently presented a significant challenge. 
Addressing this, recent work has explored the possibility of using LLMs as evaluators. While using a single LLM as an evaluation agent shows potential, it is filled with significant uncertainty and instability. 
To address these issues, we propose the \textbf{MATEval}: A ``\textbf{M}ulti-\textbf{A}gent \textbf{T}ext \textbf{Eval}uation framework"  where all agents are played by LLMs like GPT-4. 
The MATEval framework emulates human collaborative discussion methods, integrating multiple agents' interactions to evaluate open-ended text.
Our framework incorporates self-reflection and Chain-of-Thought (CoT) strategies, along with feedback mechanisms, enhancing the depth and breadth of the evaluation process and guiding discussions towards consensus, while the framework generates comprehensive evaluation reports, including error localization, error types and scoring. 
Experimental results show that our framework outperforms existing open-ended text evaluation methods and achieves the highest correlation with human evaluation, which confirms the effectiveness and advancement of our framework in addressing the uncertainties and instabilities in evaluating LLMs-generated text. Furthermore, our framework significantly improves the efficiency of text evaluation and model iteration in industrial scenarios.

\end{abstract}
\section{Introduction}

Evaluating the text generated by large language models (LLMs) has long been a challenging task, 
Traditional manual evaluation methods are not only time-consuming and laborious but also expensive\cite{human}. 
Although methods like BLEU\cite{bleu}, Rouge\cite{rouge}, and METEOR\cite{meteor} have achieved success in scenarios such as machine translation, these automated evaluation methods are limited in the context of open-ended text generation\cite{open}. 
Recently, LLMs have been used as evaluators such as G-Eval\cite{gptscore}, but these methods exhibit unstable and uncertain evaluation effects \cite{chatgpt}\cite{fair}. 
Even certain collaboration frameworks could alleviate this problem by employing multi-agent discussion, \textit{e.g.} ChatEval \cite{chateval},
however, the current methods of multi-agent collaboration remain limited to simple interactions, without fully harnessing the potential for agents' \textit{thinking} and \textit{planning}. Additionally, reaching a consensus within multi-agent discussion frameworks continues to be a challenging issue. 
Furthermore, traditional text evaluation models typically provide only a score without explaining it, making it difficult for reviewers to trust the reliability of these scores. They still need to manually identify errors, obviously slowing down the collection of bad cases and, consequently, affecting the pace of model iteration in industrial scenarios.

To address the above challenges, this paper introduces a Multi-Agent Text Evaluation Framework (MATEval). In this framework, we simulate the human collaborative process in evaluating texts generated by LLMs and propose a novel multi-agent discussion strategy. 
This strategy integrates self-reflection\cite{self} and Chain-of-Thought (CoT)\cite{cot} concepts, as self-reflection focuses on understanding the depth of issues but may lead to rigid thinking. Meanwhile, strategies based on the CoT emphasize the refinement of problems but may lack in-depth analysis of specific issues. 
Therefore, we combine the two approaches by guiding agents through prompts to decompose evaluation questions and focus on only one sub-question in each discussion round. 
During each round of the discussion, agents engage in self-reflection, considering peer inputs to enrich issue comprehension.
This approach strengthens agents' self-assessment and critical thinking, broadening their evaluation scope for open-ended text and aligning results more closely with human evaluations.

Furthermore, our framework introduces a feedback mechanism at the end of each discussion round to evaluate the quality and efficiency of the discussions, encouraging agents to reach a consensus. 
The comprehensive evaluation report generated by our framework details error types, specific locations, in-depth explanations, and corresponding scores. 

For practical applications in industry, we provide two report formats: a question-and-answer format for strategy analysis and a text report designed to help business personnel quickly identify errors and facilitate iterative improvement of LLMs. Our framework has achieved significant results in the story text evaluation task at Alipay, markedly enhancing the efficiency of model iteration.

To summarize, our main contributions in this paper are:

\begin{enumerate}

    \item We propose a Multi-Agent Evaluation Framework called MATEval\ref{framework}, which enhances the reliability of scoring by providing accurate diagnostic reports for text generated by LLMs. This framework not only facilitates model iteration in industrial scenarios but also significantly boosts audit efficiency.

    \item We propose a novel method to integrate self-reflection and CoT in our multi-agent framework. Additionally, we creatively introduce a feedback mechanism at the end of each discussion round to resolve disagreements and facilitate the achievement of consensus.

    \item We conduct comprehensive experiments on two English and two Chinese story text datasets, including one constructed based on Alipay's business story text dataset. Our experimental results showcase the effectiveness of our framework and its high correlation with human evaluations. 
    \setcounter{footnote}{0}
    \footnote{We have made the datasets and results used in our experiments publicly available at \url{https://github.com/kse-ElEvEn/MATEval}. Due to the user privacy of Alipay, we cannot make the ``Ant" dataset public.}
    
\end{enumerate}

\section{Related Work}

\textbf{Traditional NLG Evaluation}: For a significant period, open-ended text evaluation primarily depended on human annotations, which incurs substantial human and financial costs. Subsequent automated NLG evaluations employ computational models to evaluate the quality of generated texts, such as BLEU\cite{bleu}, ROUGE\cite{rouge}, and METEOR\cite{meteor}. Embedding-based metrics refer to the evaluation of generated texts by measuring the semantic similarity between generated texts and reference texts based on word or sentence embeddings. BERTScore\cite{bertscore} calculates the similarity between generated text and reference text based on BERT's contextual embedding. RUBER$_{\scriptsize \text{BERT}}$\cite{better} is also based on BERT embeddings to measure the similarity of texts with and without references through processes such as pooling and MLP operations.

\textbf{LLM-based Evaluators}: GPTScore\cite{gptscore} utilizes models such as GPT-3 to evaluate text quality, predicated on the assumption that generative pre-trained models assign higher probabilities to high-quality generated texts by given instructions and context. Recent studies also explore the potential of using ChatGPT as an NLG evaluator\cite{chatgpt}. G-Eval\cite{geval} demonstrates the evaluation of NLG outputs using prompts in LLMs like ChatGPT through Chain-of-Thought (CoT) methods.

\textbf{Communicative Agents}: Recently, the concept of using agents for communication and collaboration to accomplish specific tasks gains widespread application. CAMEL\cite{camel} introduces a cooperative agent framework called \textit{role-playing}, enabling individual agents to collaboratively solve complex tasks autonomously. ChatEval\cite{chateval} applies the multi-agent approach to text evaluation, constructing a multi-agent jury to explore the impact of different communication strategies in evaluating open-ended questions and traditional NLG tasks.

\section{Methodology}

\begin{figure}[h]
    \centering
    \includegraphics[width=0.9\linewidth]{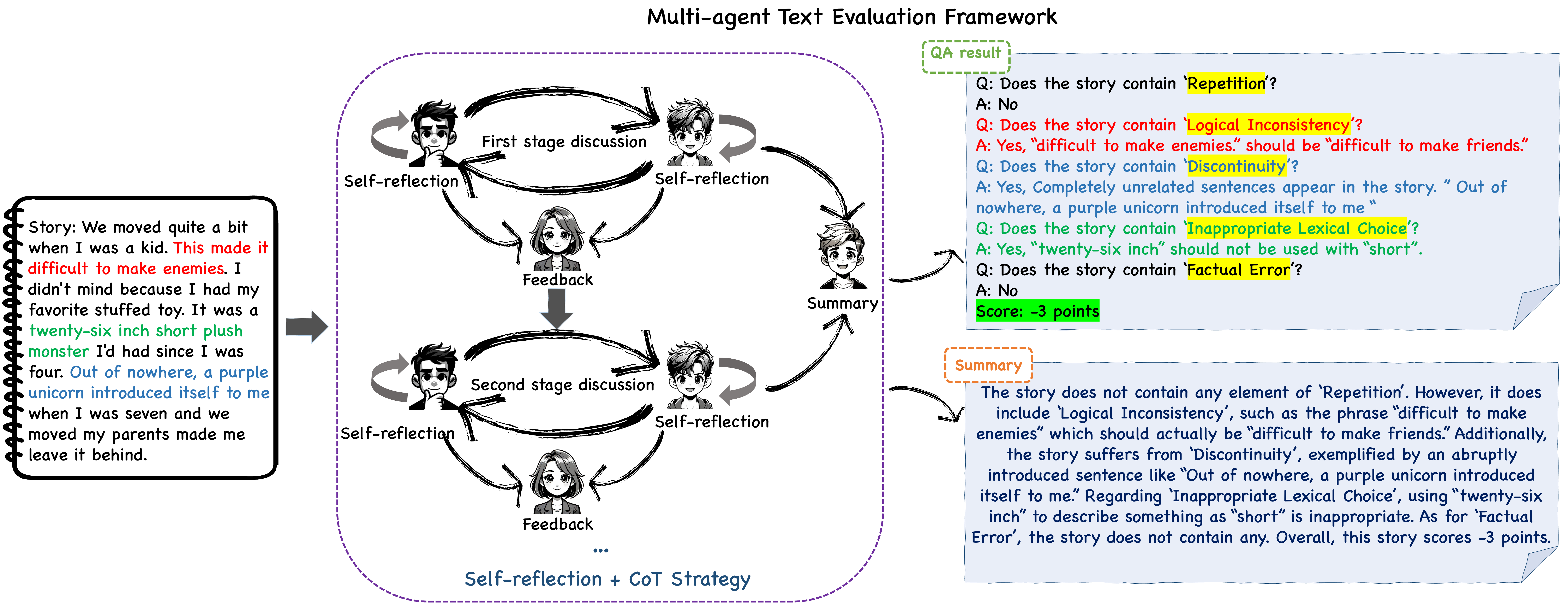}
    \caption{The overall process diagram of the MATEval Framework. The input to the framework is a text with quality questions, which after going through a multi-agent discussion that combines self-reflection and CoT strategies, outputs a detailed evaluation report.}
    \label{framework}
\end{figure}

In this section, we will provide a detailed exposition of the design of the evaluation framework within MATEval \ref{framework}, the utilization of various strategies, and the functional specifications of different roles.

\subsection{Design of the Framework}

Our framework primarily consists of agents with different roles combined with discussion strategies. The roles of agents we utilize include \textit{Evaluator Agent}, \textit{Feedback Agent}, and \textit{Summarizer Agent}, who collaborate to complete text evaluation tasks. The \textit{Evaluator Agent} is the main entity in the evaluation task, the \textit{Feedback Agent} plays a crucial role in improving discussion quality and promoting consensus, and the \textit{Summarizer Agent} is indispensable for consolidating discussion information, summarizing, and forming evaluation reports. In our framework, we employ a discussion strategy that integrates self-reflection and CoT.

The framework accepts text as input, which may contain various quality issues. The output is a detailed evaluation report outlining error type, location, explanation, and score. We present the results in two formats: one is a Q\&A format conducive to evaluating correlation, allowing easy extraction of correlation scores for similarity calculations. The other is a report format that is conducive to iteration by relevant business personnel. This enables business personnel to quickly identify text issues and refine models using the analysis reports, enhancing efficiency. This is shown in the right part of Figure \ref{framework}.

\subsection{Application of Different Roles}

In this section, we introduce several key roles within our framework and their respective functions. 

\begin{figure}[t]
    \centering
    \includegraphics[width=1\linewidth]{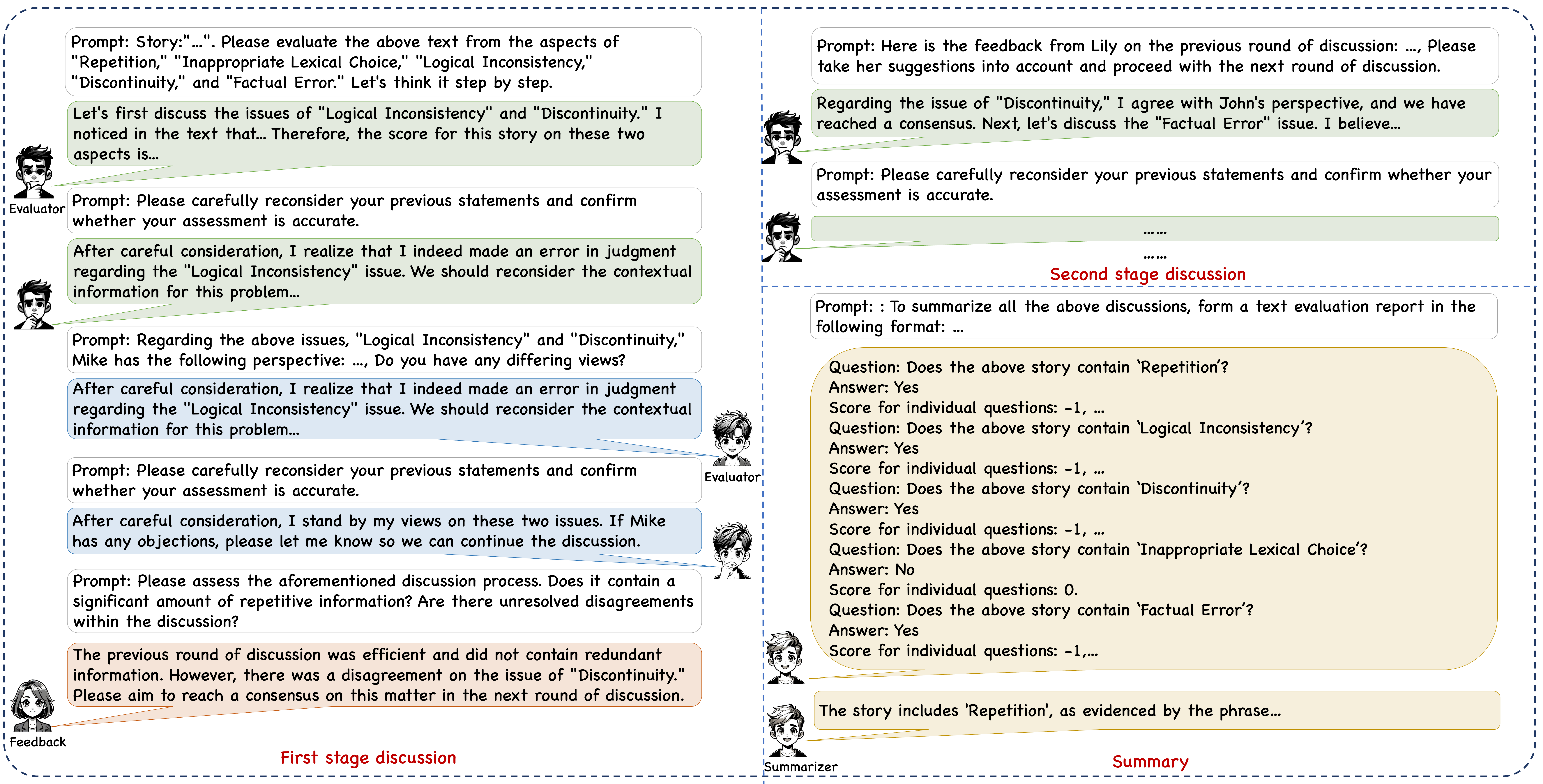}
    \caption{The diagram includes prompts and dialogue that incorporates a process of discussion with self-reflection, CoT, feedback mechanisms and final summary.}
    \label{chat-flow}
\end{figure}

\textbf{\textit{Evaluator Agent:}} The core element in the framework is the evaluator, for which we use GPT-4 to conduct multi-round evaluations and responses that are guided through carefully designed prompts. The evaluator stores and processes statements from other agents, using this as a reference for dialogue history. Each agent not only receives responses from others but also provides their own statements, with the entire process requiring minimal human intervention.

\textbf{\textit{Feedback Agent}}:  The feedback agent evaluates the content and quality of each discussion round. It focuses on identifying inefficient dialogues and disagreements. If issues are detected, it suggests improvements for subsequent rounds to enhance efficiency and consensus through prompts.

\textbf{\textit{Summarizer Agent}}: After all discussions are concluded, the summarizer compiles the entire process and outcomes. It provides a Q\&A format evaluation report, detailing the identification, analysis, and scoring of various issues. 
Additionally, we provide a comprehensive text-based format evaluation report that includes detailed problem descriptions and is easy to read to help improve model performance in industrial production.

\subsection{Feedback Mechanism}
The feedback mechanism is a well-designed component in our framework. At the end of each discussion round, we use a prompt to guide a \textit{feedback agent} to summarize and evaluate the discussion. Its role is to steer subsequent discussions towards less repetition, enhance the efficiency of the discussion, and importantly, guide the participants towards reaching a consensus. All of these are achieved by conveying the feedback provider's remarks to the agents involved in the discussion.

\subsection{Combined Self-reflection and CoT}

\textbf{\textit{Self-reflection Strategy}}: After each agent's statements, they engage in a process of self-reflection. Guided by the prompt, agents adjust their statements by integrating the viewpoints of other agents. The final statements of each agent are stored and used as historical information for subsequent discussions. In a new round of discussion, the statements from the previous round are stored as historical information.

\textbf{\textit{CoT Strategy}}: We guide agents through prompts to autonomously decompose problems and address only one sub-problem in each round of discussion. Meanwhile, each agent's statements are stored and used as historical information for subsequent discussions.

\textbf{\textit{Combined Self-reflection and CoT}}: As shown in Algorithm \hyperref[alg:text_eval_SR_CoT]{1}. Combining self-reflection and the CoT is an important strategy employed in our framework. Agents autonomously decompose the question according to the prompt, focusing on one sub-question in each round of multiple discussions: $\mathcal{D}(\mathcal{Q}) = \{\mathcal{Q}_\text{1}, \mathcal{Q}_\text{2}, \ldots, \mathcal{Q}_\text{n}\}$, $\mathcal{P}_\text{i} = \mathcal{I}(\mathcal{Q}_\text{i}, \mathcal{H})$ , where $\mathcal{I}$ represents the formation of preliminary ideas based on the prompt, $\mathcal{Q}$ is the evaluation task, $\mathcal{Q}_\text{n}$ is the sub-question and $\mathcal{H}$ is the history information. They then optimize their statements through self-reflection: $\mathcal{R}_\text{i} = \mathcal{S}(\mathcal{P}_\text{i}, \mathcal{H})$. Next, update the history: $\mathcal{H} = \mathcal{H} \cup \{\mathcal{R}_i\}$.  After each round of discussion, a feedback provider evaluates the discussion to reduce repetition and disagreement: $\mathcal{E} = \text{E}(\mathcal{H})$ . Finally, a summarizer compiles all statements to produce the evaluation report: $\mathcal{R}_{\text{final}} = \text{Summary}(\mathcal{H})$. The overall prompt and the flow of the discussion are illustrated in Figure \ref{chat-flow}.

\begin{algorithm}
\caption{Self-Reflection and Chain-of-Thought Discussion with Feedback}
\label{alg:text_eval_SR_CoT}
\begin{algorithmic}[1]
\Input Given text, set of sub-questions $\mathcal{Q}$ decomposed by LLMs-based agents, number of agents $\mathcal{N}$
\Output Final evaluation report $R_{\text{final}}$
\State $\mathcal{H} \gets \{\}$, Agents $\gets \{\mathcal{A}_1, \mathcal{A}_2, \ldots, \mathcal{A}_N\}$, SubQIndex $\gets 0$ 
\While{SubQIndex $<$ length($\mathcal{Q}$)}
    \State SubQCurrent $\gets \mathcal{Q}[\text{SubQIndex}]$ \Comment{select the current sub-question}
    \State SubQIndex $\gets \text{SubQIndex} + 1$ \Comment{increment the sub-question index}
    \For{each $\mathcal{A}_i$ in Agents}
        \State $\mathcal{P}_i \gets \text{Formulate\_Idea}(\mathcal{Q}_\text{i}, \mathcal{H})$ \Comment{generate preliminary ideas}
        \State $\mathcal{R}_i \gets \text{Self\_Reflection}(\mathcal{P}_i, \mathcal{H})$ \Comment{reflect on their statements}
        \State $\mathcal{H} \gets \mathcal{H} \cup \{\mathcal{R}_i\}$ \Comment{update historical record with reflected statement}
    \EndFor
    \State $\mathcal{E} \gets \text{Evaluate}(\mathcal{H})$ \Comment{evaluator assesses this round's discussion}
    \State $\mathcal{F}_{\text{round}} \gets \text{Feedback}(\mathcal{E})$  \Comment{provide feedback $F_{\text{round}}$ to all agents}
\EndWhile
\State $\mathcal{R}_{\text{final}} \gets \text{Summary}(\mathcal{H})$ \Comment{summarizer generates error analysis report}
\State \Return $\mathcal{R}_{\text{final}}$
\end{algorithmic}
\end{algorithm}

\section{Experiments}
\subsection{Implementation Details}

In the MATEval framework, we select OpenAI's GPT-4 as our LLMs due to its outstanding performance and API accessibility. We set the temperature parameter to 0 for result reproducibility. GPT-4's easy access facilitated effective and coherent multi-agent interactions in our experiments.

\subsection{Dataset}

We mainly apply our framework to the evaluation of story texts generated by LLMs in Alipay business scenarios. So in the experiment, we mainly focus on two open-ended story datasets: ROCStories (\textbf{ROC})\cite{openmeva} and WritingPrompts (\textbf{WP})\cite{openmeva}.
Considering GPT-4's context length limitations and the need for storing multi-round discussion contexts, we truncate WP stories to the first 200 words, ensuring textual integrity at the truncation point. To test model generalizability, we conduct similar experiments on two Chinese datasets. These include the \textit{Chinese \textbf{LO}ng \textbf{T}ext understanding and generation} \textbf{(LOT)}\cite{lot} dataset, comprising human-written stories averaging 106 words, and a dataset of Chinese fairy tales constructed using prompts from Alipay's business data with GPT-3.5 named \textbf{Ant}, mainly involving fables and fairy tales with an average story length of 125 words. Considering GPT-4's request rate limits and high usage costs, we select the first 200 stories from each dataset for multi-agent discussion experiments.

Using the GPT-4 interface, we introduce five basic types of errors into 200 story texts across different datasets to simulate possible problems in stories. They are \textbf{\textit{Repetition}\textbf{(REP)}, \textit{Logical Inconsistency} \textbf{(LINC)}, \textit{Discontinuity} \textbf{(DCONT)}, \textit{Inappropriate Lexical Choice }\textbf{(ILC)}, and \textit{Factual Error} \textbf{(FER)}}. Repetition includes redundant sentences or excessive word use; Logical Inconsistency encompasses antonym substitution and polarity shifts in sentences; Discontinuity involves sequencing errors or unrelated content; Inappropriate Lexical Choice refers to misused quantifiers or pronouns; and Factual Error denotes contradictions with established knowledge. We hire five annotators to assess these datasets, ensuring the data aligns with human preferences.
Both manual and multi-agent scoring follow the same criteria: starting from zero, each error deducts one point, with scores tallied for each error type and the total for each text.

\subsection{Compared Methods}

\textbf{Referenced Metrics}:  The \textbf{\textit{BLEU}}\cite{bleu} score is used to evaluate lexical similarity between candidate and reference texts. \textbf{\textit{ROUGE-L}}\cite{rouge} focuses on the longest common subsequence to assess the fluency and coherence of texts. And \textbf{\textit{RUBER-BERT}\cite{better}, }an enhancement of the original RUBER model with BERT's contextual embeddings, includes both referenced and unreferenced versions. The referenced version, \textbf{\textit{RUBER-BERT{\small r}}}, measures the similarity between candidate responses and reference texts using BERT word embeddings.

\textbf{Unreferenced Metrics}:  The RUBER's unreferenced version \textbf{\textit{RUBER-BERT{\small u}}}\cite{better} predicts relevance between responses and queries using BERT word embeddings followed by operations such as pooling and MLP. The BERT-based \textbf{\textit{UNION}}\cite{union} model distinguishes human-written stories from automatically generated negative samples and corrects the interference of negative samples. \textbf{\textit{ChatEval}}\cite{chateval} evaluates open-ended Q\&A quality through multi-agent framework, and we select its most effective \textit{One-by-One} approach for comparison.

\textbf{Our Methods}: In our experience, we compare various strategies:
\begin{itemize}
    \item \textbf{Single-Agent}\textbf{(SA)}: LLMs directly evaluate stories without multi-agent.
    \item \textbf{One-by-One}\cite{chateval}\textbf{(O\_b\_O)}: Agents sequentially evaluate stories in multi-round discussions, without optimization strategies.
    \item \textbf{Self-Reflection (SR)}: Agents conduct self-reflection, considering their and others' previous statements during discussions.
    \item \textbf{Chain-of-Thought (CoT)}: Agents break down the assessment problem through prompts, solving one sub-problem in each discussion round.
    \item \textbf{Self-Reflection + CoT (SR+CoT)}: By combining CoT and self-reflection strategies, agents first decompose questions for discussion, then engage in self-reflection each round.
\end{itemize}

In all these strategies, we employ feedback mechanisms at the end of each discussion round, as well as a final summary.

\subsection{Experimental Results}

Our experiments on the ROC and WP datasets are presented in Table \ref{ROC}, using MATEval framework strategies to evaluate narrative texts. We calculate Spearman ($\rho$) and Kendall ($\tau$) correlation coefficients to compare models' evaluations with human judgments.

\begin{table}[t]
\begin{center}
    \caption{Correlation of evaluation results with human judgment using different models and different strategies of MATEval on the ROC/WP dataset, where SA stands for Single-Agent, SR denotes Self-Reflection, and CoT represents Chain-of-Thought. The symbols $\rho$/$\tau$ respectively indicate the Spearman/Kendall correlation. The highest correlation values are highlighted in bold.}
    \label{ROC}
    \scalebox{1.0}{
        \begin{tabular}{llcccccccccc}
 
        \toprule

        \multicolumn{2}{c}{\multirow{2}{*}{Strategy}} & \multicolumn{2}{c}{REP} & \multicolumn{2}{c}{LINC} & \multicolumn{2}{c}{DCONT} & \multicolumn{2}{c}{ILC} & \multicolumn{2}{c}{FER} \\
        \cmidrule(lr){3-4} \cmidrule(lr){5-6} \cmidrule(lr){7-8} \cmidrule(lr){9-10} \cmidrule(lr){11-12}
        && $\rho$ & $\tau$ & $\rho$ & $\tau$ & $\rho$ & $\tau$ & $\rho$ & $\tau$ & $\rho$ & $\tau$ \\
 
        \midrule

        \multirow{10}{*}{ROC} & BLEU & 0.318 & 0.260 & 0.193 & 0.153 & 0.156 & 0.128 & 0.037 & 0.031 & -0.010 & -0.008 \\
        & ROUGE-$_{\scriptsize \text{L}}$ & -0.017 & -0.014 & 0.129 & 0.102 & 0.202 & 0.165 & 0.056 & 0.045 & 0.104 & 0.084 \\
        & RUBER$_{\scriptsize \text{r}}$ & 0.036 & 0.035 & 0.054 & 0.049 & 0.315 & 0.297 & -0.018 & -0.017 & -0.176 & -0.166 \\
        & RUBER$_{\scriptsize \text{u}}$ & -0.111& -0.091& 0.038& 0.031& 0.131& 0.107& 0.134& 0.110& 0.180& 0.146\\ 
        & UNION& -0.093& -0.076& 0.091& 0.071& -0.018& -0.015& 0.057& 0.046& 0.072&0.059\\ 
        & SA & 0.699 & 0.694 & 0.268 & 0.253 & 0.318 & 0.312 & 0.240 & 0.236 & 0.545 & 0.538 \\ 
        & O\_b\_O & 0.698 & 0.692 & 0.170 & 0.160 & 0.356 & 0.349 & 0.259 & 0.248 & 0.484 & 0.473 \\ 
        & SR& 0.691 & 0.680 & 0.169 & 0.154 & 0.354 & 0.339 & 0.144 & 0.138 & 0.498 & 0.478 \\
        & CoT & 0.743 & \textbf{0.737} & 0.189 & 0.180 & 0.288 & 0.282 & 0.213 & 0.205 & 0.502 & 0.491 \\ 
        & SR+CoT& \textbf{0.735} & 0.728 & \textbf{0.281} & \textbf{0.264} & \textbf{0.391} & \textbf{0.382} & \textbf{0.263} & \textbf{0.256} & \textbf{0.575} & \textbf{0.561} \\

        \midrule

        \multirow{10}{*}{WP} & BLEU & 0.087 & 0.071 & 0.096 & 0.073 & 0.039 & 0.033 & -0.114 & -0.091 & 0.009 & 0.007 \\
        & ROUGE-$_{\scriptsize \text{L}}$ & 0.092 & 0.074 & 0.127 & 0.096 & 0.083 & 0.068 & -0.046 & -0.037 & 0.049 & 0.040 \\
        & RUBER$_{\scriptsize \text{r}}$ & 0.038 & 0.036 & -0.020 & -0.018 & -0.081 & -0.076 & 0.035 & 0.033 & 0.076 & 0.071 \\
        & RUBER$_{\scriptsize \text{u}}$ & -0.102& -0.084& 0.054& 0.041& -0.006& -0.005& -0.006& -0.007& 0.111&0.089\\
        & UNION& 0.048& 0.039& 0.010& 0.008& -0.110& -0.090& -0.038& -0.031& 0.052&0.042\\
        & SA & 0.258& 0.246& 0.107& 0.095& 0.111& 0.105& 0.192& 0.1802& 0.176&0.171\\ 
        & O\_b\_O & 0.386& 0.380& 0.183& 0.166& 0.081& 0.075& 0.089& 0.082& \textbf{0.299}&\textbf{0.286}\\ 
        & SR& \textbf{0.491}& \textbf{0.483}& 0.120& 0.107& 0.224& 0.209& 0.057& 0.051& 0.214&0.208\\ 
        & CoT & 0.132& 0.129& 0.159& 0.139& 0.203& 0.191& 0.002& 0.001& 0.218&0.211\\
        & SR+CoT& 0.430& 0.417& \textbf{0.215}& \textbf{0.188}& \textbf{0.265}& \textbf{ 0.248}& \textbf{ 0.290}& \textbf{0.266}& 0.299&0.286\\

        \bottomrule

        \end{tabular}
    }
\end{center}
\end{table}

Analyzing the experimental results on the ROC and WP datasets, we draw the following conclusions:

LLMs-based methods show better Spearman ($\rho$) and Kendall ($\tau$) correlations compared to traditional n-gram and Bert-based methods, proving their effectiveness in text evaluation.

Multi-agent discussions generally surpass single-agent evaluations in performance, suggesting they significantly improve text evaluation quality.

In analyzing the strategies used within the MATEval framework, we found that employing self-reflection or the CoT independently produces unstable results across different error types. In some cases, these methods even underperformed compared to single-agent evaluations. This might be due to inherent flaws when applying these strategies separately. For example, self-reflection can lead to rigid thinking in multiple discussion rounds, where agents often repeat earlier content without adding new insights. On the other hand, using CoT alone often results in superficial and divergent perspectives, offering only a basic analysis of each error type without delving deeper.
    
The combination of self-reflection and CoT achieved the best overall correlation, particularly excelling over other methods in evaluating \textit{Logical Inconsistency}, \textit{Discontinuity}, and \textit{Inappropriate Lexical Choice}. It significantly improves the evaluation of \textit{Discontinuity} compared to the single-agent method, demonstrating the framework's high sensitivity to textual coherence. However, its effectiveness was lower for \textit{Repetition} and \textit{Factual Error}. Agents often misidentified emotionally similar sentences as repetitive, despite clear definitions. Furthermore, the framework's evaluation of \textit{Factual Errors} was limited by LLMs constraints, specifically the absence of external knowledge affecting common sense error detection, highlighting the need to integrate external knowledge for future multi-agent framework enhancements.

\begin{table}[t]
\begin{center}
    \caption{Correlation of evaluation results with human judgment using different models and different strategies of MATEval on the LOT/Ant dataset. The highest correlation values are highlighted in bold.}
    \label{LOT}
    \scalebox{1.0}{
        \begin{tabular}{llcccccccccc}
        \toprule
        \multicolumn{2}{c}{\multirow{2}{*}{Strategy}} & \multicolumn{2}{c}{REP} & \multicolumn{2}{c}{LINC} & \multicolumn{2}{c}{DCONT} & \multicolumn{2}{c}{ILC} & \multicolumn{2}{c}{FER} \\
        \cmidrule(lr){3-4} \cmidrule(lr){5-6} \cmidrule(lr){7-8} \cmidrule(lr){9-10} \cmidrule(lr){11-12}
        && $\rho$ & $\tau$ & $\rho$ & $\tau$ & $\rho$ & $\tau$ & $\rho$ & $\tau$ & $\rho$ & $\tau$ \\
        \midrule
        \multirow{5}{*}{LOT} 
        & SA & \textbf{0.829}& \textbf{0.817}& 0.120& 0.110& 0.336& 0.324& 0.179& 0.175& 0.284& 0.279\\
        & O\_b\_O & 0.770& 0.764& 0.142& 0.131& 0.249& 0.239& 0.069& 0.066&\textbf{ 0.362}& \textbf{0.349}\\
        & SR& 0.751& 0.735& 0.054& 0.048& 0.282& 0.267& 0.118& 0.112& 0.296& 0.284\\
        & CoT & 0.636& 0.628& 0.026& 0.024& 0.215& 0.206& 0.051& 0.049& 0.155& 0.151\\
        & SR+CoT& 0.811& 0.798&\textbf{ 0.197}& \textbf{0.185}& \textbf{0.354}&\textbf{0.341}&\textbf{0.182}& \textbf{0.175}& 0.341& 0.333\\

        \midrule

        \multirow{5}{*}{Ant} 
        & SA & 0.522& 0.517& 0.281& 0.275& 0.231& 0.231& 0.318& 0.316& 0.495&0.489\\ 
        & O\_b\_O & 0.545& 0.538& 0.145& 0.141& -0.010& -0.010& 0.011& 0.018& 0.347&0.343\\ 
        & SR& 0.563& 0.557& 0.185& 0.175& 0.069& 0.069& 0.187& 0.184& 0.368&0.360\\ 
        & CoT & 0.572& 0.562& 0.034& 0.039& 0.024& 0.024& 0.040& 0.042& 0.358&0.352\\
        & SR+CoT& \textbf{0.694}& \textbf{0.676}& \textbf{0.403}& \textbf{0.377}& \textbf{0.287}& \textbf{0.282}& \textbf{0.424}&\textbf{ 0.417}& \textbf{0.552}&\textbf{0.544}\\

        \bottomrule
        \end{tabular}
    }
\end{center}
\end{table}

\begin{table}[t]
\begin{center}
    \caption{Correlation of evaluation results with human judgment of ablation experiments on the ROC dataset.}
    \label{ROC_ablation}
    \scalebox{1.0}{
        \begin{tabular}{llcccccccccc}
        \toprule
        \multicolumn{2}{c}{\multirow{2}{*}{Strategy}} & \multicolumn{2}{c}{REP} & \multicolumn{2}{c}{LINC} & \multicolumn{2}{c}{DCONT} & \multicolumn{2}{c}{ILC} & \multicolumn{2}{c}{FER} \\
        \cmidrule(lr){3-4} \cmidrule(lr){5-6} \cmidrule(lr){7-8} \cmidrule(lr){9-10} \cmidrule(lr){11-12}
        && $\rho$ & $\tau$ & $\rho$ & $\tau$ & $\rho$ & $\tau$ & $\rho$ & $\tau$ & $\rho$ & $\tau$ \\
        \midrule
        & MATEval{\scriptsize \text{-FB}} & 0.567& 0.567& 0.039& 0.038& 0.259& 0.255& 0.266& 0.260&0.477& 0.473\\
        
        & MATEval{\scriptsize \text{-QA}}& 0.612& 0.597& 0.071& 0.068& 0.011& 0.011& 0.132& 0.127& 0.283& 0.281\\
        
        & MATEval{\scriptsize \text{-multi}}& 0.699 & 0.694 & 0.268 & 0.253 & 0.318 & 0.312 & 0.240 & 0.236 & 0.545 & 0.538 \\ 
        
        & MATEval& \textbf{0.735} & \textbf{0.728} & \textbf{0.281} & \textbf{0.264} & \textbf{0.391} & \textbf{0.382} & \textbf{0.263} & \textbf{0.256} & \textbf{0.575} & \textbf{0.561} \\
        \bottomrule
        \end{tabular}
    }
\end{center}
\end{table}

\subsection{Ablation Study}

To assess the effectiveness of different modules in MATEval, we conducted ablation experiments on the ROC dataset, involving the removal of the feedback mechanism, omitting Q\&A format explanations, and not using a multi-agent approach. Table \ref{ROC_ablation} demonstrates that the complete MATEval framework surpasses its ablated versions, confirming the significance of both feedback mechanisms, explanations and multi-agent methods. Specifically, this establishes the importance of our feedback mechanisms in promoting discussion consensus and enhancing relevance. It also proves that providing scoring explanations by LLMs significantly enhances evaluation effectiveness.

\subsection{Generalization Experiments}

To verify the generalizability of our framework across different languages and in the industrial field, we experimented with two Chinese datasets: \textit{LOT} and a story text dataset \textit{Ant}, derived from Alipay's business data. The findings in Table \ref{LOT} were similar to those from English datasets. Interestingly, single-agent evaluations often perform better in Chinese, possibly due to its unique language and sentence structure. This suggests that optimizing agents for different languages may require tailored adjustments for optimal performance.

\begin{figure}[h]
    \centering
    \includegraphics[width=1\linewidth]{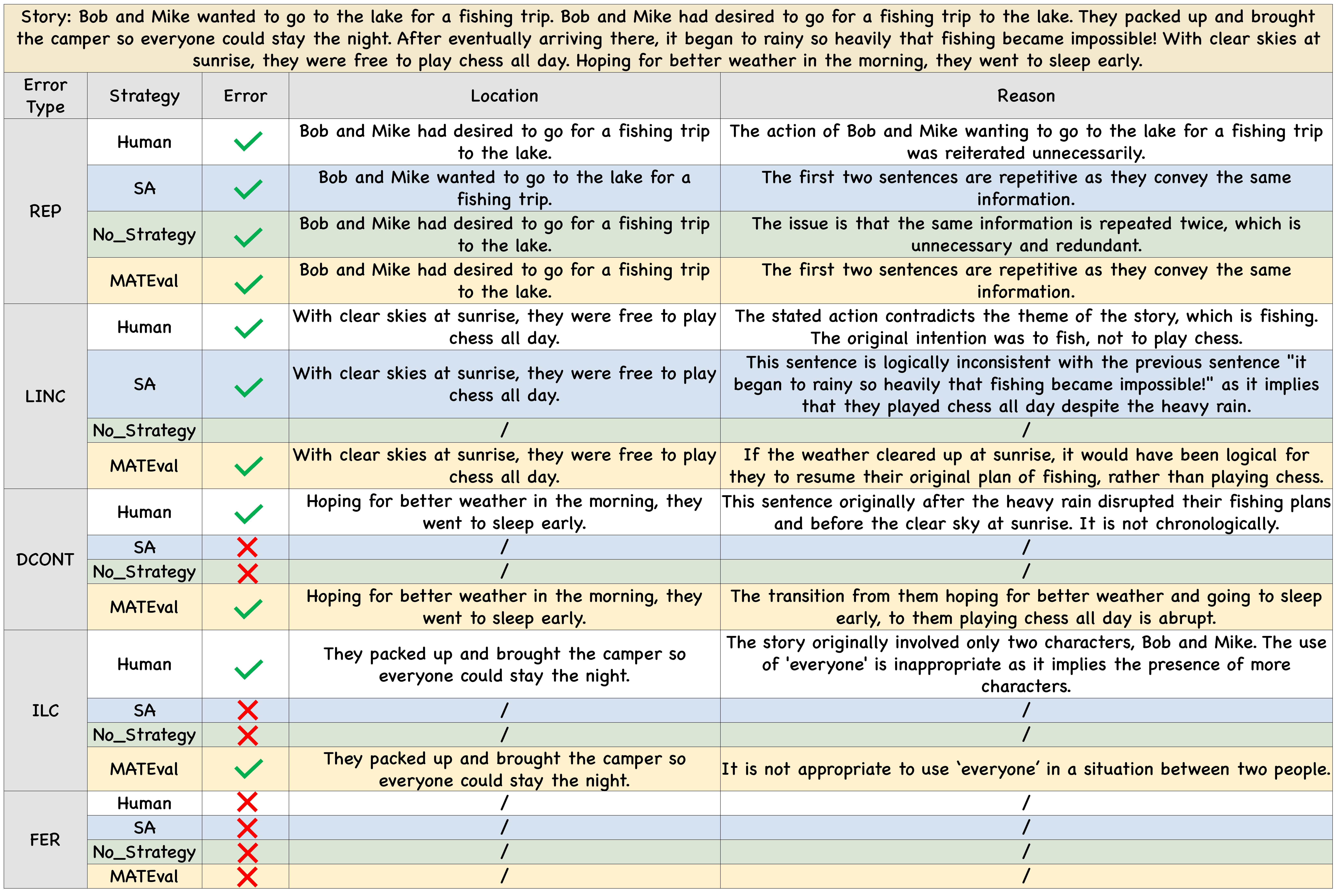}
    \caption{The schematic diagram illustrating the comparison of results generated by the MATEval framework and other methods.}
    \label{case}
\end{figure}

\subsection{Case Study}

To showcase the MATEval framework's effectiveness in real-world industrial settings, we illustrated its application using a story text example shown in Table \ref{case}. We compared our framework against manual evaluation, single-agent evaluation, and the strategy-less One\_by\_One evaluation method. Experimental results demonstrate that our method is basically consistent with human evaluations, unlike other methods, which show some gaps, thereby confirming the high correlation between our approach and human evaluation.

\section{Conclusion}

In this paper, we proposed the MATEval framework, which enhances the evaluation performance of open-ended story text generated by LLMs in the industrial field. Extensive experiments show that MATEval’s evaluation results on two classic story datasets are more aligned with human preferences than those of existing methods. In the Alipay industrial scenario, our framework significantly improves review efficiency and evaluation accuracy, serving as an effective aid.\footnote{This work was supported by Ant Group}\footnote{This work was supported by the Natural Science Foundation of China (Grant No. U21A20488). We thank the Big Data Computing Center of Southeast University for providing the facility support on the numerical calculations in this paper.}

In future work, we can fine-tune LLMs as agents in the industrial field to complete specific domain tasks. When solving a complex domain task, we can enable these domain-specific agents to collaborate with each other, thereby enhancing their capability and efficiency in addressing challenges.

%
%
%
\bibliographystyle{splncs04}
\bibliography{reference}
\end{document}